\documentclass{article}

\usepackage{microtype}
\usepackage{graphicx}
\usepackage{subfigure}
\usepackage{booktabs} % for professional tables

\usepackage[utf8]{inputenc} % allow utf-8 input
\usepackage[T1]{fontenc}    % use 8-bit T1 fonts
\usepackage{url}            % simple URL typesetting
\usepackage{booktabs}       % professional-quality tables
\usepackage{amsfonts}       % blackboard math symbols
\usepackage{nicefrac}       % compact symbols for 1/2, etc.
\usepackage{microtype}      % microtypography
\usepackage{floatrow}
\newfloatcommand{capbtabbox}{table}[][\FBwidth]

\usepackage[table,xcdraw]{xcolor}
\usepackage{multirow}
\usepackage{graphicx}
\usepackage{verbatim} % for multiline comments
\usepackage{amsfonts}
\usepackage{placeins}

\usepackage[accepted]{icml2019}
\usepackage{float}
\floatstyle{plaintop}
\restylefloat{table}
\usepackage{titlesec}
\makeatletter
\@addtoreset{section}{part}
\makeatother
\titleformat{\part}[display]
{\LARGE\bfseries}{}{0pt}{}
\definecolor{myblue}{rgb}{0,0.08,0.45}
\usepackage{hyperref}       % hyperlinks
\hypersetup{
    colorlinks,
    citecolor=myblue,
    filecolor=myblue,
    linkcolor=myblue,
    urlcolor=myblue
}
\icmltitlerunning{Towards Explainable Anticancer Compound Sensitivity Prediction via Multimodal Attention-based Convolutional Encoders}

\begin{document}

\twocolumn[
\icmltitle{Towards Explainable Anticancer Compound Sensitivity Prediction via Multimodal Attention-based Convolutional Encoders}



\icmlsetsymbol{equal}{*}

\begin{icmlauthorlist}
\icmlauthor{Matteo Manica}{equal,ibm}
\icmlauthor{Ali Oskooei}{equal,ibm}
\icmlauthor{Jannis Born}{equal,ibm,eth,usz}
\icmlauthor{Vigneshwari Subramanian}{aac}
\icmlauthor{Julio Sáez-Rodríguez}{aac,hei}
\icmlauthor{María Rodríguez Martínez}{ibm}
\end{icmlauthorlist}

\icmlaffiliation{ibm}{IBM Research Zurich, Switzerland}
\icmlaffiliation{eth}{ETH Zurich, Switzerland}
\icmlaffiliation{usz}{University of Zurich, Switzerland}
\icmlaffiliation{aac}{RWTH Aachen University, Germany}
\icmlaffiliation{hei}{Heidelberg University, Germany}

\icmlcorrespondingauthor{Matteo Manica, Ali Oskooei, Jannis Born}{\{tte,osk,jab\}@zurich.ibm.com}

%\icmlcorrespondingauthor{Matteo Manica}{tte@zurich.ibm.com}
%\icmlcorrespondingauthor{Ali Oskooei}{osk@zurich.ibm.com}
%\icmlcorrespondingauthor{Jannis Born}{jab@zurich.ibm.com}

\icmlkeywords{Multiscale, Multimodal, Attention, CNN, RNN, Explainability, Interpretability, Molecular networks, Molecular fingerprints, GDSC, SMILES, Gene expression, Drug discovery, Drug sensitivity, Anticancer compounds, IC50, EC50, Lead discovery, Personalized medicine, Precision medicine}

\vskip 0.3in
]



\printAffiliationsAndNotice{\icmlEqualContribution} % otherwise use the standard text.

\begin{abstract}

In line with recent advances in neural drug design and sensitivity prediction, we propose a novel architecture for interpretable prediction of anticancer compound sensitivity using a multimodal attention-based convolutional encoder.
Our model is based on the three key pillars of drug sensitivity: compounds' structure in the form of a SMILES sequence, gene expression profiles of tumors and prior knowledge on intracellular interactions from protein-protein interaction networks.
We demonstrate that our multiscale convolutional attention-based (MCA) encoder significantly outperforms a baseline model trained on Morgan fingerprints, a selection of encoders based on SMILES as well as previously reported state of the art for multimodal drug sensitivity prediction (R2 = 0.86 and RMSE = 0.89). %0.96).
Moreover, the explainability of our approach is demonstrated by a thorough analysis of the attention weights. We show that the attended genes significantly enrich apoptotic processes and that the drug attention is strongly correlated with a standard chemical structure similarity index.
Finally, we report a case study of two receptor tyrosine kinase (RTK) inhibitors acting on a leukemia cell line, showcasing the ability of the model to focus on informative genes and submolecular regions of the two compounds.
The demonstrated generalizability and the interpretability of our model testify its potential for in-silico prediction of anticancer compound efficacy on unseen cancer cells, positioning it as a valid solution for the development of personalized therapies as well as for the evaluation of candidate compounds in de novo drug design.

\end{abstract}

\section{Introduction}
\label{seq:intro}

\subsection{Motivation}
Discovery of novel compounds with a desired efficacy and improving existing therapies are key bottlenecks in the pharmaceutical industry, which fuel the largest R\&D business spending of any industry and account for 19\% of the total R\&D spending worldwide~\cite{petrova2014innovation, goh2017smiles2vec}.
Anticancer compounds, in particular, take the lion’s share of drug discovery R\&D efforts, with over 34\% of all drugs in the global R\&D pipeline in 2018 (5,212 of 15,267 drugs)~\citep{lloyd2017pharma}.
Despite enormous scientific and technological advances in recent years, serendipity still plays a major role in anticancer drug discovery~\citep{hargrave2012serendipity} without a systematic way to accumulate and leverage years of R\&D to achieve higher success rates in drug discovery.
On the other hand, there is strong evidence that the response to anticancer therapy is highly dependent on the tumor genomic and transcriptomic makeup, resulting in heterogeneity in patient clinical response to anticancer drugs~\cite{geeleher2016cancer}.
This varied clinical response has led to the promise of personalized (or precision) medicine in cancer, where molecular biomarkers, e.g., the expression of specific genes, obtained from a patient’s tumor profiling may be used to choose a personalized therapy.

These challenges highlight a need across both pharmaceutical and healthcare industries for multimodal quantitative methods that can jointly exploit disparate sources of knowledge with the goal of characterizing the link between the molecular structure of compounds, the genetic and epigenetic alterations of the biological samples and drug response~\citep{de2016algorithms}.
In this work, we present a multimodal attention-based convolutional encoder that enables us to tackle the aforementioned challenges. 

\subsection{Related work}
There have been a plethora of works on the prediction of drug sensitivity in cancer cells~\cite{garnett2012systematic, yang2012genomics, costello2014community, ali2018machine,kalamara2018find}. While the majority of them have focused on the analysis of unimodal datasets (genomics or transcriptomics, e.g.,~\citet{de2016algorithms,tan2016prediction,turki2017link,tan2018drug}), a handful of previous works have integrated omics and chemical descriptors to predict cell line-drug sensitivity using a variety of methods including but not limited to: simple neural networks (one hidden layer) and random forests~\citep{menden2013machine}, kernelized Bayesian matrix factorization~\citep{ammad2014integrative}, Pearson correlation-based similarity networks~\citep{zhang2015predicting}, a Kronecker product kernel in conjunction with SVMs~\citep{wang2016inferences}, autoencoders in combination with elastic net and SVMs~\citep{ding2018precision}, matrix factorization~\citep{wang2017improved}, trace norm regularization~\citep{yuan2016multitask}, link predictions~\citep{stanfield2017drug} and collaborative filtering~\citep{liu2018anti, zhang2018hybrid}.
In addition to genomic and chemical features, previous studies have demonstrated the value of complementing drug sensitivity prediction models with prior knowledge in the form of protein-protein interactions (PPI) networks~\citep{oskooei2018network}.
For example, in a network-based per-drug approach integrating these data sources,~\citet{zhang2018novel} surpassed various earlier models and reported a performance drop of 3.6\% when excluding PPI information. 

However, all previous attempts at incorporating chemical information in drug sensitivity prediction rely on molecular fingerprints as chemical descriptors.
Traditionally, fingerprints were applied extensively for drug discovery, virtual screening and compound similarity search~\cite{cereto2015molecular}, but it has recently been argued that the usage of engineered features constraints the learning ability of machine learning algorithms~\citep{goh2017smiles2vec}.
Furthermore, for many applications, molecular fingerprints may not be relevant, informative or even available.

With the rise of deep learning methods and their proven ability to learn the most informative features from raw data, machine learning methods used in molecular design and drug discovery have also experienced a shift~\citep{chen2018rise,wu2018moleculenet,grapov2018rise}.
For instance, computational chemists borrowed methods from neural language models~\cite{bahdanau2014neural} to encode SMILES strings of molecules and predict chemical properties of molecules~\cite{jastrzkebski2016learning, goh2017smiles2vec, schwaller2018molecular}. 
Once a gold standard in sequence modeling, recurrent neural networks were initially employed as SMILES encoders~\citep{goh2017smiles2vec,bjerrum2017smiles,segler2017generating}.
However, it has been recently shown that convolutional architectures are superior to RNNs for sequence modeling~\citep{bai2018empirical}, and specifically for modeling the SMILES string encoding of compounds~\citep{kimber2018synergy}.
It is noteworthy that these findings are in agreement with our model comparison results that reveal convolutional architectures are superior for SMILES sequence modeling.

Most recently,~\citet{chang2018cancer} adopted deep learning methods to develop a pan-drug model for predicting IC50 drug sensitivity of drug-cell line pairs.
Utilizing > 30,000 binary features ($\sim$3,000 for the molecular drug fingerprint and the rest for a genomic fingerprint), they employ a model ensemble of 5 deep convolutional networks (4 are completely linear) with convolutions applied separately to each of the genomic and molecular features before the encodings are merged.
While working towards a common goal, our approaches are vastly different.
Our method presents several key advantages: First, our algorithm ingests raw information (SMILES string representation) which in turn enables data augmentation and boosts model performance~\citep{bjerrum2017smiles}. Secondly, we apply convolutions only on SMILES representations for which convolutions are meaningful (i.e., convolutions combine information from various molecular substructures).
In accordance with~\citet{costello2014community}, we use transcriptomic features (gene expression profiles) instead of genomic features. Moreover, we combine transcriptomic and molecular information using a contextual attention encoder that renders our model transparent and interpretable, a feature that is paramount in precision medicine and has only recently started to be tackled~\citep{yang2018linking}.
An additional key advantage of our approach is our strict splitting strategy and evaluation criterion. While previous works relied on lenient splitting strategies that ensured no drug-cell line~\textit{pair} in the test data was seen during training, we adopt a more stringent splitting strategy and deprive the model training of all drugs and cell lines that are present in the test dataset.
Our strict training and evaluation strategy results in a significantly more challenging problem but in turn ensures the model is learning generalizable molecular substructures with anticancer properties as opposed to memorizing drug sensitivity from cell-drug pairs that it has encountered during training.
A model that has been trained with such a criterion will generalize better to completely unseen drugs and cell lines thus paving the way for both, in silico validation of de novo drug candidates in pharmaceutics and selection of a suitable therapy in personalized medicine.
A lenient split, on the other hand, may facilitate drug repositioning, as it performs best when drug and cell line have been encountered during training.
\begin{figure}[!htb]
\centering
\includegraphics[width=\linewidth]{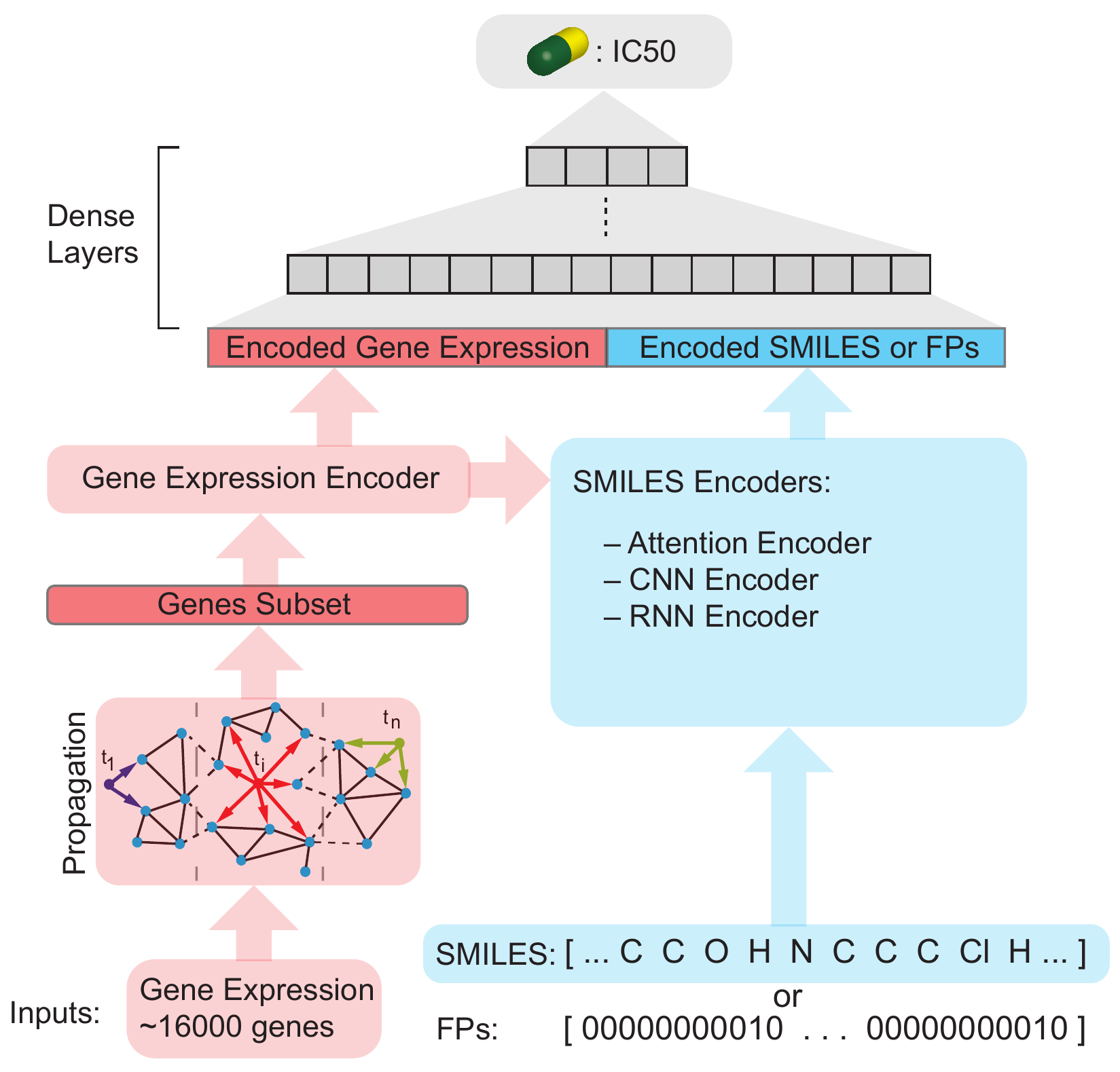}%
\caption{\textbf{The multimodal end-to-end architecture of the proposed encoders.} General framework for the explored architectures. Each model ingests a cell-compound pair and makes an IC50 drug sensitivity prediction. Cells are represented by the gene expression values of a subset of 2,128 genes, selected according to a network propagation procedure. Compounds are represented by their SMILES string (apart from the baseline model that uses 512-bit fingerprints). The gene-vector is fed into an attention-based gene encoder that assigns higher weights to the most informative genes. To encode the SMILES strings, several neural architectures are compared and used in combination with the gene expression encoder in order to predict drug sensitivity.}
\label{fig:architectures}
\end{figure}
\subsection{Scope of the presented work}
In this work we build upon our previous work on multimodal drug sensitivity prediction using attention-based encoders~\cite{oskooei2018paccmann}, and propose a novel best-performing architecture, an attention-based multiscale convolutional encoder. In addition, we perform a thorough validation of the attention weights given by our proposed MCA model. We combine 1) cell line data, 2) molecular structure of compounds and 3) prior knowledge of protein interactions to predict drug sensitivity. Specifically, for 1) we explore the usage of gene expression profiles and for 2) we explore different neural architectures in combination with our devised contextual attention architecture to encode raw SMILES of anticancer drugs in the context of the cell that they are acting on (see \autoref{fig:architectures}).
We show that attention-based SMILES encoders significantly surpass a baseline feedforward model utilizing Morgan (circular) fingerprints~\citep{rogers_extended-connectivity_2010} ($p$ < 1e-6 on RMSE).
Using our multiscale convolutional attentive (MCA) encoder, we show that we achieve superior IC50 prediction performance on the GDSC database ~\citep{iorio2016landscape} compared with the existing methods~\citep{menden2013machine,chang2018cancer}.
Utilizing SMILES representations is highly desirable, as they are ubiquitously available and more interpretable than traditional fingerprints.
Furthermore, our contextual attention mechanism emerges as the key component of our proposed SMILES encoder, as it helps validate our findings by explaining the model's inner working and reasoning process, many of which are in agreement with domain-knowledge on biochemistry of cancer cells.

\section{Methods}
\label{seq:methods}

\subsection{Data}
\label{sseq:data}
Throughout this work, we employed drug sensitivity data from the the publicly available Genomics of Drug Sensitivity in Cancer (GDSC) database~\cite{iorio2016landscape}.
The database includes the screening results of more than a thousand genetically profiled human pan-cancer cell lines with a wide range of anticancer compounds (both chemotherapeutic drugs and targeted therapeutics).
The drug sensitivity values were represented by half maximal inhibitory concentration (IC50, i.e., the micromolar concentration of a drug necessary to inhibit 50\% of the cells) on the log-scale.
We focused on targeted drugs with publicly available molecular structure (208 compounds of 265 in total) and retrieved the molecular structure of the compounds from \texttt{\href{https://pubchem.ncbi.nlm.nih.gov/search/}{PubChem}}~\citep{kim2018pubchem} and the \texttt{\href{http://lincs.hms.harvard.edu/db/}{LINCS}} database.
From the collected canonical SMILES, Morgan fingerprints were acquired using \texttt{\href{http://www.rdkit.org}{RDKit}} (512-bit with radius 2).
Exploiting the property that most molecules have multiple valid SMILES strings, the data augmentation strategy proposed by~\citet{bjerrum2017smiles} was adopted to represent each anticancer compound with 32 different SMILES strings. We chose to represent each cell by its transcriptomic profile as it has been demonstrated that transcriptomic data are more predictive of drug sensitivity when compared to other omic data~\cite{costello2014community}. As such, all available RMA-normalized gene expression data were retrieved from the GDSC database resulting in transcriptomic profiles of 985 cell lines in total.

\subsection{Network propagation}
\label{sseq:propagation}
Each of the 985 cell lines was initially represented by the expression levels of 17,737 genes which we then reduced to a subset of 2128 genes through network propagation over the STRING protein-protein interaction (PPI) network~\citep{szklarczyk2014string}, a comprehensive PPI database including interactions from multiple data sources.
Following the procedure described in~\citet{oskooei2018network}, STRING was used to incorporate intracellular interactions in our model by adopting a network propagation scheme for each drug, where the weights associated with each of the reported targets were diffused over the STRING network (including interactions from all the evidence types) leading to an importance distribution over the genes (i.e., the vertices of the network).
Our adopted weighting and network propagation scheme consisted of the following steps: we first assigned a high weight ($W$ = 1) to the reported drug target genes while assigning a very small positive weight ($\varepsilon=1\mathrm{e}{-5}$) to all other genes.
Thereafter, the initialized weights were propagated over STRING.
This process was meant to integrate prior knowledge about molecular interactions into our weighting scheme, and simulate the propagation of perturbations within the cell following the drug administration.
Let us denote the initial weights as $W_{0}$ and the string network as $S = (P, E, A)$, where $P$ are the protein vertices of the network, $E$ are the edges between the proteins and $A$ is the weighted adjacency matrix.
The smoothed weights are determined from an iterative solution of the propagation function~\cite{oskooei2018network}:
\begin{equation}
\label{eq:propagation}
W_{t+1} =\alpha W_{t}A' + (1-\alpha)W_{0}
\end{equation}
where $D$ is the degree matrix and  $A'$ is the normalized adjacency matrix, obtained from the degree matrix $D$:
\begin{equation}
A'=D^{-\frac{1}{2}}AD^{-\frac{1}{2}}
\end{equation}
The diffusion tuning parameter, $\alpha$ ($0\leq\alpha\leq1$), defines how far the prior knowledge weights can diffuse through the network.
In this work, we used $\alpha=0.7$, as recommended in the literature for the STRING network~\citep{hofree2013network}.
Adopting a convergence rule of $e=(W_{t+1}-W_t) < 1 \mathrm{e}{-6}$, we solved~\autoref{eq:propagation} iteratively for each drug and used the resultant weights distribution to determine the top $20$ highly ranked genes for each drug.
By selecting the top 20 genes for every drug, it was possible to compile an interaction-aware subset of genes (2,128 genes in total).
This subset containing the most informative genes was then used to profile each cell line in the dataset before it was fed into our models.
We limited the selection to the top 20 genes for every drug to guarantee a trade-off between topology-awareness and the number of features describing the biomolecular profile.
We then paired all screened cell lines and drugs to generate a pan-drug dataset of cell-drug pairs and the associated IC50 drug response. Due to missing values in the GDSC database, pairing of the 985 cell lines with the 208 drugs resulted in 175,603 pairs which could be augmented to more than 5.5 million data points following SMILES augmentation~\citep{bjerrum2017smiles}.

\subsection{Model architectures}
The majority of previous efforts in drug sensitivity prediction focused on traditional molecular descriptors (fingerprints). Morgan fingerprints have been shown to be a highly informative representation for many chemical prediction tasks~\cite{unterthiner2014deep,chang2018cancer}.
We explored several neural network SMILES encoder architectures to investigate whether the molecular information of compounds, in the context of drug sensitivity prediction, can be learned directly from the raw SMILES rather than using engineered fingerprints. As such, all explored encoder architectures were compared against a baseline model that utilized 512-bit Morgan fingerprints. The general architecture of our models is shown in \autoref{fig:architectures}.

\textbf{Deep baseline (DNN).}
The baseline model is a 6-layered DNN with [512, 256, 128, 64, 32, 16] units and a sigmoid activation. The hyperparameters for the baseline model were optimized via a cross-validation scheme (see \autoref{sseq:data_split}). 512-bit Morgan fingerprints and gene expression profiles (filtered using the network propagation described in~\autoref{sseq:data}) were concatenated into a joint representation from the first layer onwards.

\begin{figure}[!htb]
\centering
\includegraphics[width=\linewidth]{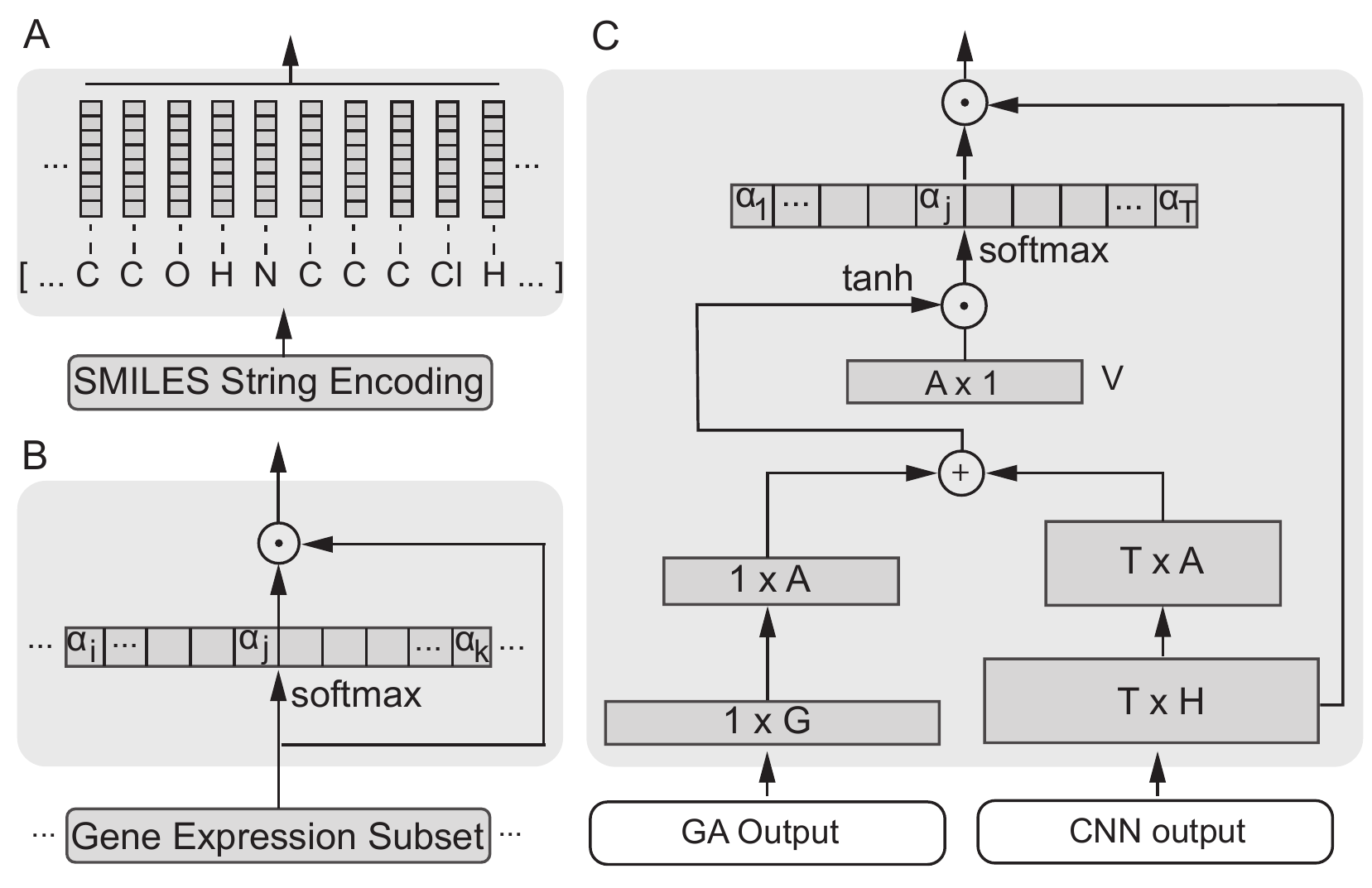}%
\caption{\textbf{Key layers employed throughout the SMILES encoder.} A) An embedding layer transforms raw SMILES strings into a sequence of vectors in an embedding space. B) An attention-based gene expression encoder generates attention weights that are in turn applied to the input gene subset via a dot product.  C) A contextual attention layer ingests the SMILES encoding (either raw or the output of another encoder, e.g., CNN, RNN and so on) of a compound and genes from a cell to compute an attention distribution ($\alpha_{i}$) over all tokens of the SMILES encoding, in the context of the genetic profile of the cell. The attention-filtered molecule represents the most informative molecular substructures for IC50 prediction, given the gene expression of a cell.}
\label{fig:layers}
\end{figure}

\textbf{SMILES models (commonalities).}
To investigate which model architecture best learns the molecular information of compounds, we explored various SMILES encoders.
All SMILES-based models ingest the expression profiles and the SMILES text encodings for the structure of the compounds.
The SMILES sequences were tokenized using a regular expression~\cite{schwaller2018found}. 
%This ensured that small functional units of the molecule (such as \texttt{[NH]} or \texttt{[N+]}) were represented as single tokens to the model. 
The resulting atomic sequences were zero-padded and represented as $E=\{e_1,..., e_T\}$, with learned embedding vectors $e_i \in \mathbb{R}^{H}$ for each dictionary token (see \autoref{fig:layers}\textcolor{myblue}{A}).
Each cell line, represented by the genetic subset selected through network propagation, is fed to the gene attention encoder (see \autoref{fig:layers}\textcolor{myblue}{B}).
A single dense softmax layer with the same dimensionality as the input produces an attention weight distribution over the genes and filters them in a dot product, ensuring most informative genes are given a higher weight for further processing.
The resulting gene attention weights render the model interpretable, as they identify genes that drive the sensitivity prediction for each cell line.
This architecture was also investigated for the deep baseline model but discarded due to inferior performance.
All SMILES encoders were followed by a set of dense layers (as shown in \autoref{fig:architectures}) with dropout ($p_{drop}$ =  0.5) for regularization and sigmoid activation function.
The regression was completed by a single neuron with linear activation (rather than sigmoid) to avoid restricting the values between 0 and 1 and hinder the learning process of the network as a result.
\begin{figure}[!htb]
\centering
\includegraphics[width=\linewidth]{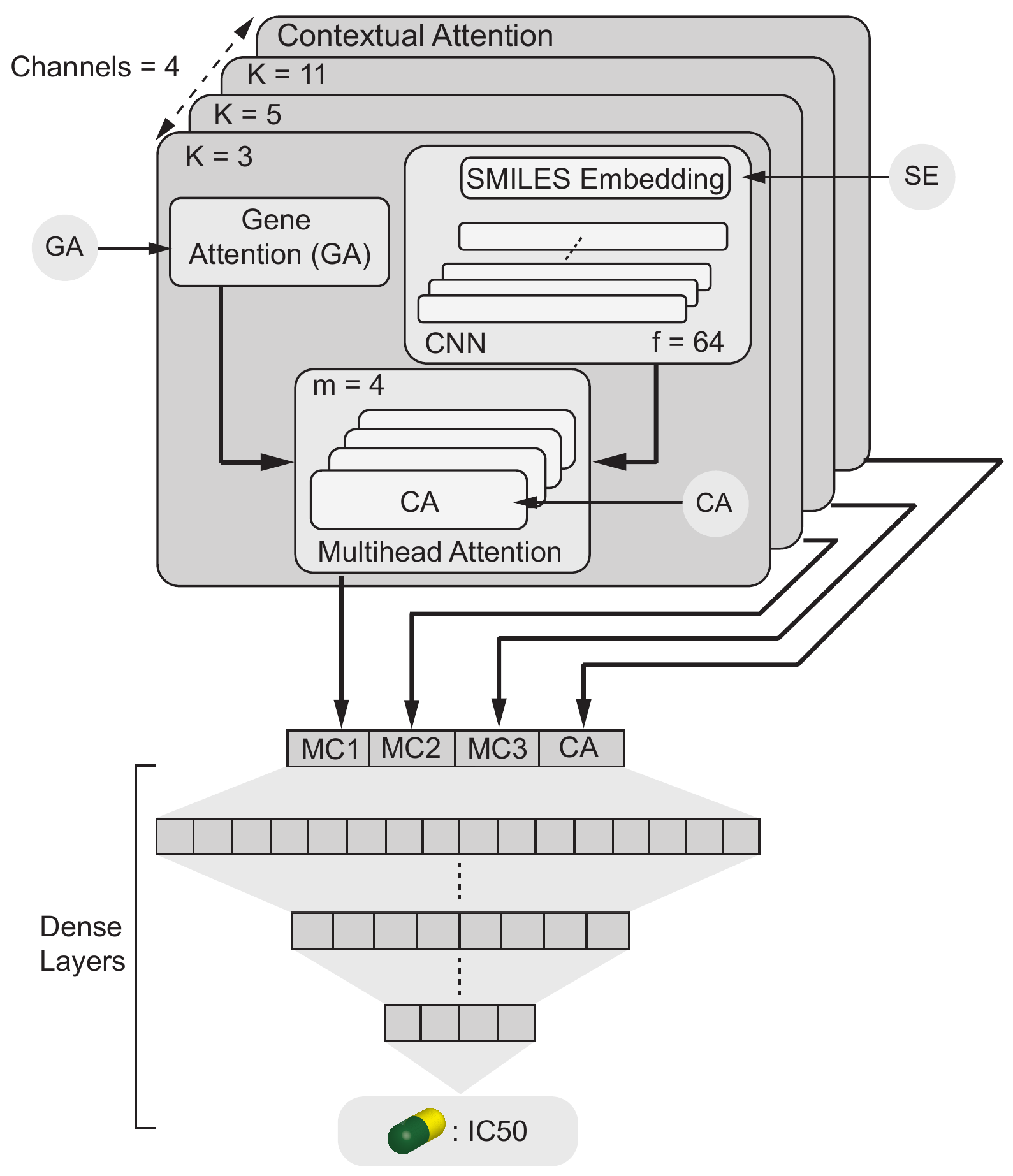}%
\caption{\textbf{Model architecture of the multiscale convolutional attentive (MCA) encoder.} The MCA model employed 3 parallel channels of convolutions over the SMILES sequence with kernel sizes $K$ and one residual channel operating directly on the token level. Each channel applied a separate gene attention layer, before (convoluted) SMILES and filtered genes were fed to a multihead of 4 contextual attention layers. The output of these 16 layers were concatenated and resulted in an IC50 prediction through a stack of dense layers. For CA, GA and SE, see~\autoref{fig:layers}.}
\label{fig:mca}
\end{figure}

\textbf{Bidirectional recurrent (bRNN).}
RNNs have traditionally been the first-line approach for sequence encoding. To investigate their effectiveness in encoding SMILES, we adopted a 2-layered bidirectional recurrent neural network (bRNN) with gated recurrent units (GRU)~\cite{cho2014learning}.
The final states of the forward and backward GRU-RNN were concatenated and fed to the dense layers for IC50 prediction.

\textbf{Stacked convolutional encoder (SCNN).}
Next, we employed an encoder with four layers of stacked convolutions and sigmoid activation function.
2D convolution kernels in the first layer collapsed the embedding vectors' hidden dimensionality while subsequent 1D convolutions extracted increasingly long-range dependencies between different parts of the molecule.
As a result, similarly to the bRNN, any output neuron of the SCNN SMILES encoder had integrated information from the entire molecule.

\textbf{Self-attention (SA).}
We investigated several encoders that leveraged neural attention mechanisms, originally introduced by~\citet{bahdanau2014neural}.
Interpretability is paramount in healthcare and drug discovery ~\citep{koprowski2018machine}.
As such, neural attention mechanisms are central in our models as they enable us to explain and interpret the observed results in the context of underlying biological and chemical processes.
Our first attention configuration is a self-attention (SA) mechanism adapted from document classification~\citep{yang2016hierarchical} for encoding SMILES strings.
The SMILES attention weights $\alpha_{i}$ were computed per atomic token as:
\begin{equation}
\label{eq:seq_att}
\alpha_{i} = \frac{\exp(u_{i})}{\sum_{j}^{T} \exp(u_{j}) } \hspace{2mm} \textmd{where} \hspace{2mm} u_{i} = V^{T} \tanh(W_{e}s_{i}+b) 
\end{equation}
The matrix $W_{e}\in \mathbb{R}^{A\times H}$ and the bias vector $b \in \mathbb{R}^{A\times 1}$ are learned in a dense layer.
$s_{i}$ is an encoding of the $i$-th token of the molecule, in the most basic case simply the SMILES embedding $e_{i}$.
In all attention mechanisms, the encoded smiles are obtained by filtering the inputs with the attention weights.

\textbf{Contextual-attention (CA).}
Alternatively, we devised a contextual attention (CA) mechanism that utilizes the gene expression subset $G$ as a context (\autoref{fig:layers}\textcolor{myblue}{C}). The attention weights $\alpha_{i}$ are determined according to the following equation :
\begin{equation}
\label{eq:contex_att}
u_{i} = V^{T} \tanh(W_{e}s_{i} + W_{g}G)  \hspace{2mm} \textmd{where} \hspace{2mm}   W_{g} \in \mathbb{R}^{A\times |G|}
\end{equation}
First, the matrices $W_{g}$ and $W_{e}$ project both genes $G$ and the encoded SMILES tokens $s_{i}$ into a common attention space, $A$.
Adding the gene context vector to the projected token ultimately yields an $\alpha_{i}$ that denotes the relevance of a compound substructure for drug sensitivity prediction, given a gene subset $G$.

\textbf{Multiscale convolutional attention (MCA).} In their simplest form, the attention mechanisms of the SA and CA model operates directly on the embeddings, disregarding positional information and long-range dependencies. Instead they exploit the frequency counts on individual tokens (atoms, bonds). Interestingly, the attention models nevertheless outperform the bRNN and SCNN which integrated information from the entire molecule.
In order to combine the benefits of the attention-based models, i.e., interpretability with the ability of sequence encoders to extract both local and long-range dependencies, we devised the multiscale convolutional attentive (MCA) encoder shown in \autoref{fig:mca}.
Using MCA, the SMILES string of a compound is analyzed using three separate channels, each convolving the SMILES embeddings with a set of $f$ kernels of sizes [$H$, 3], [$H$, 5] and [$H$, 11] and ReLU activation.
The efficacy of a drug may be tied primarily to the occurrence of a specific molecular substructure. MCA is designed to capture substructures of various size using its variable kernel size. For instance, a particular kernel could detect a steroid structure, typical across anticancer molecules~\citep{gupta2013current}.
Following the multiscale convolutions, the resulting feature maps of each channel were fed into a contextual attention layer that received the filtered genes as context.
Similarly to~\citet{vaswani2017attention}, we employed $m$ = 4 contextual attention layers for each channel, in order to allow the model to jointly attend several parts of the molecule.
The multi-head attention approach, counteracts the tendency of the softmax to filter out the vast majority of the sequence steps~\citep{lifeature}.
In a fourth channel, the convolutions were skipped (residual connection) and the raw SMILES embeddings were directly fed to the parallel CA layers.
The output of these 4$m$ layers were concatenated before given to the stack of dense feedforward layers.

\subsection{Model evaluation}
\label{sseq:data_split}

\textbf{Strict split.}
To benchmark the different proposed architectures, a strict data split approach was adopted to ensure neither the cell lines nor the compound structures within the validation or test datasets have been seen by our models prior to validation or testing. 
This is in contrast to previously published pan-drug models which have explored only a lenient splitting strategy, where both compound and cell-line of any sample in the test dataset were encountered during training.
In our data split strategy, 10\% subsets of the total number of 208 compounds and 985 cell lines from the GDSC database were set aside to be used as an unseen test dataset to evaluate the trained models.
The remaining 90\% of compounds and cell lines were then used in a 25-fold cross-validation scheme for model training and validation.
In each fold, 4\% of the drugs and 4\% of cell lines were separated and used to generate the validation dataset and the remaining drugs and cell lines were paired and fed to the model for training. In practice, this strategy deprived the model from a significant proportion of samples which were not sorted into any of training, validation or testing data.
We decided to choose 25-fold cross-validation because: 1) this number is large enough to employ tests of statistical significance across different models and 2) To increase the size of the training set and in turn improve the performance of the trained models by decreasing the number of pairs that were excluded from training set (i.e., the validation set).

\textbf{Lenient split.}
To compare our model with prior works that chose a less strict data split strategy, we adopted a similar strategy that rather than depriving the model from both the cells and drugs in the test set, ensured no cell-drug pair in the test set has been seen before. The new split consisted of a standard 5-fold cross-validation scheme, wherein 10\% of the pairs (175,603 pairs from 985 cell lines and 208 drugs) were set aside for testing.
IC50 values of the training data were normalized to $[0,1]$ and the same transformation was applied to validation and test data. Gene expression values in the training set were standardized and the same transformation was applied to the gene expression in the validation and test sets.

\subsection{Training procedure}
\label{sseq:training}
All described architectures were implemented in \texttt{TensorFlow 1.10} with a MSE loss function that was optimized with Adam ($\beta_{1}=0.9$, $\beta_{2}=0.999$, $\varepsilon=1\mathrm{e}{-8}$) and a decreasing learning rate~\cite{kingma2014adam}. An embedding dimensionality of $H$ = 16 was adopted for all SMILES encoders. The attention dimensionality was set to $A$ = 256 for the SA and CA model, while $A$ = $f$ = 64 for MCA.
In the final dense layers of all models we employed dropout ($p_{drop}$ = 0.5), batch normalization and a sigmoid activation.
All models were trained with a batch size of 2,048 for a maximum of 500k steps on a cluster equipped with \texttt{POWER8} processors and an \texttt{NVIDIA Tesla P100}.
\section{Results}
\label{seq:results}

\subsection{Model performance comparison on strict split}
\label{ssec:stringent_split}
\autoref{tab:performance} compares the test performance of all models trained using a 25-fold cross-validation scheme. 
%% NEW TABLE
\begin{table}[!htb]
\centering
\caption{
\textbf{Performance of the explored architectures on test data following 25-fold cross-validation}. The median Root Mean Square Error (RMSE) and the Interquartile Range (IQR) between predicted and true IC50 values on test data of all 25 folds are reported. Interestingly, attention-based models outperform all other models including models trained on fingerprints with a statistically significant margin (* indicating a significance of $p$ < 0.01 compared to the DNN encoder, ** to the MCA). 
}
\label{tab:performance}
\scalebox{0.65}{
\begin{tabular}{ccccccc}
\toprule \toprule
\multirow{2}{*}{\bfseries Encoder type} & {\bfseries Drug} &
 {\bfseries Standardized RMSE}
 \\ 
& \bfseries structure & Median $\pm$ IQR \\  \midrule
\textbf{Deep baseline (DNN)} & Fingerprints &  0.122 $\pm$ 0.010 \\ \midrule
\textbf{Bidirectional recurrent (bRNN)} & SMILES &  0.119 $\pm$ 0.011  \\ \midrule
\textbf{Stacked convolutional (SCNN)} & SMILES &  0.130 $\pm$ 0.006 \\ \midrule
\textbf{Self-attention (SA)} & SMILES & 0.112* $\pm$ 0.009 \\ \midrule
\textbf{Contextual attention (CA)} & SMILES & 0.110* $\pm$ 0.007 \\ \midrule
\textbf{Multiscale convolutional attentive (MCA)} & SMILES & 0.109* $\pm$ 0.009 \\ \midrule 
\textbf{MCA (prediction averaging)} & SMILES & \bfseries 0.104** $\pm$ 0.005 \\ \midrule
\end{tabular}}
\end{table}
As shown in \autoref{tab:performance}, the MCA model yielded the best performance in predicting drug sensitivity (IC50) of unseen drugs-cell line pairs within the test dataset. Since IC50 was normalized to [0,1], the observed RMSE implies an average deviation of 10.4\% of the predicted IC50 values from the true values.
Interestingly, the bRNN SMILES encoder matched, but did not surpass the performance of the baseline model (DNN).
The SCNN encoder which combined and encoded information from across the entire SMILES sequence, performed significantly worse than the baseline, as assessed by a one-sided Mann-Whitney-U test ($U$ = 126, $p$ < 2e-4).
We therefore hypothesize that local features of the SMILES sequence (such as counts of atoms and bonds) contain information most predictive of a drug's efficacy.
Attention-based models that operated directly on the SMILES embeddings (SA, CA), performed significantly better than all previous models (e.g., CA vs. DNN: $U$ = 42, $p$ < 9e-8, SA vs. DNN: $U$ = 82, $p$ < 5e-6).
Surprisingly, neither complementing the SMILES embedding with positional encodings (similarly to \citet{vaswani2017attention}) nor complementing the bRNN encoder with attention was found to improve the model performance.
Ultimately, the MCA model was devised to combine token-level information (benefical for the attention-only models) with spatially more holistic chemical features within the same model. 
By architecture, some convolution kernels in the MCA could for example develop a sensitivity for a pyrimidine ring, potentially indicative of a tyrosine kinase inhibitor (such as Gefitinib, Afatanib or Erlotinib); an enzyme which inhibits phosphorylation of epidermal growth factor receptors (EGFR) to suppress tumor cell proliferation~\citep{jiao2018advances}. 
The MCA model also outperformed the baseline model significantly ($U$ = 136, $p$ < 3e-4) like an improved version employing prediction averaging across the 20 best checkpoints did ($U$ = 10, $p$ < 2e-9 and $U$ = 152, $p$ < 9e-4 comparing to the plain MCA).
In general, we observed a strong variability across the folds, leading us to report median as a more robust measure of performance than the mean across the folds. The variability across the folds stemmed from the strict splitting strategy (see \autoref{sseq:data_split}) that resulted in training, validation and test datasets that were significantly different from one another.
In conclusion, our results suggest that in order to effectively capture the mode of action of a compound, we require information from a combination of token-level (i.e., atom or bond level) and longer range dependencies across the SMILES sequence.

%All presented SMILES-based neural architectures are a combination of 1) a drug-cell line encoder and 2) a regressor mapping the encoding to drug sensitivity.
%An assumption therein is that a set of dense layers trained jointly with the encoder is the best regressor to map the drug-cell line encoding onto an IC50 value.
%In a case study, we extracted the activations of a fully trained MCA before the dense layers (MC1-MC4 in \autoref{fig:mca}) and fitted standard regression models (SVM, Random Forest, AdaBoost, Gradient boosting) to these encodings (see Supplementary Material \autoref{tab:regression_models}). The results indicate that a joint training of a drug-cell line encoder with some dense layers for regression is superior to fitting regressors offline to the same encodings.\\
%As we have described in~\autoref{sseq:data_split}, the main challenge of predicting drug sensitivity lies in well-capturing the effect of compounds (as opposed to an informative encoding of the cancer type).

\subsection{Model validation on lenient data split}
\label{ssec:lenient_split}
In addition to the performance evaluations in \autoref{tab:performance}, we evaluated the MCA model using a less strict data split strategy comparable to what had been adopted in previous works~\citep{chang2018cancer}. This allowed for a more meaningful comparison between the performance of our models with existing prior art.
\begin{figure}
\includegraphics[width=1\linewidth]{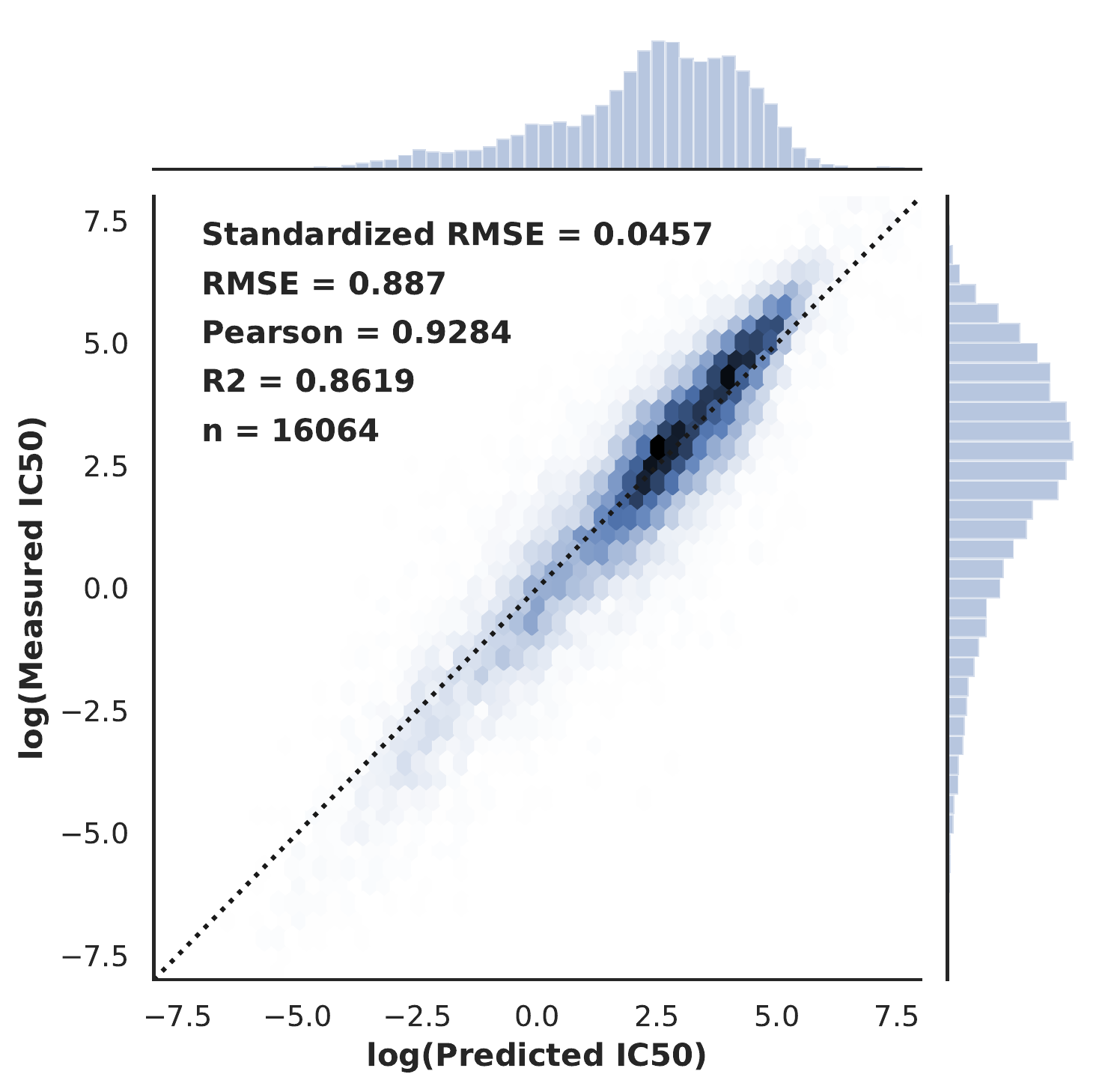}%
    \caption{\textbf{Test performance of MCA on lenient splitting.} Scatter plot of correlation between true and predicted drug sensitivity by a late-fusion model ensemble of all five folds. RMSE refers to the plotted range, since the model was fitted in log space. }
\label{fig:lenient_scatter}
\end{figure}
As \autoref{fig:lenient_scatter} shows, the MCA model ensemble achieved a RMSE of 0.887 on the log-IC50-scale, corresponding to a deviation of 4.6\%. 
The explained variance of 86.19\% suggests that our model learned to a significant extent to predict the IC50 value of an unknown pair, when both the cell line and drugs in the test set were not excluded from the training set. In addition, we observed that MCA's performance on the lenient split was on par or superior to that of existing pan-drug models~\citep{menden2013machine,chang2018cancer}.
%The CDRscan model reported by~\citet{chang2018cancer} achieved a RMSE of 1.07 and a R2 of 0.84. 
%although they used more than 10 times as many input features than we do (SMILES: 155, cell line: 2,128). 
%Sample sizes were comparable, CDRscan employed engineered fingerprints from \texttt{PaDEL} whereas our MCA model ingested raw SMILES strings without augmentation (augmentation has been shown to improve model performance in~\citet{bjerrum2017smiles}) to ensure the training dataset size for both models are roughly the same.
\subsection{Attention analysis}

\textbf{Drug structure attention.}
To quantify and analyze the drug attention on a large scale, we retrieved attention profiles for a panel of drug-cell line pairs where each drug has been evaluated for all the cell lines in the set. The selected panel consisted of 150 drugs and 200 cell lines.
For each drug, we defined a matrix of pairwise Euclidean distances between the attention profiles of the treated cell lines.
The resulting distance matrix quantifies the variation in attention profiles of a drug as a function of the treated cell lines.
We then computed, for each pair of drugs, the Frobenius distance between the attention distance matrices defined above.
Finally, we evaluated the correlation between the Frobenius distances of each pair of drugs and their Tanimoto coefficient \citep{tanimoto1958elementary}, an established index for evaluating drug similarity based on fingerprints \citep{bajusz2015tanimoto}.
This approach resulted in a Pearson correlation of $\rho$ =0.64 ($n$ = 22500, $p$ < 1e-50).
The fact that the attention similarity of any two drugs is highly correlated with their structural similarity indicates that the model indeed learns valuable insights on structural properties of compounds.

\textbf{Gene attention.}
In order to thoroughly validate the gene attention weights, we computed the attention profiles of all cell lines in the test data, averaged the
attention weights and filtered them by discarding genes with negligible attention values ($a_i < \frac{1}{K}$, where $K$ is the number of genes in the panel).
Based on the resulting subset of 371 highly attended genes, we performed a pathway enrichment analysis using Enrichr~\cite{chen2013enrichr, kuleshov2016enrichr}. The goal was to identify relevant processes highlighted by the genes the model learned to focus on.
The analysis revealed a significant activation (adjusted $p$ < 0.004) of the apoptosis signaling pathway in PANTHER~\cite{mi2016panther}.
In essence, drug sensitivity prediction is connected to apoptosis and our analysis suggests that the model learns to focus on genes associated with key molecular processes elicited by anticancer compounds, e.g., programmed cellular death.

\textbf{A case study: two TK inhibitors.}
As a further validation, we analyzed in detail the neural attention mechanism of the best MCA model (lenient split) for two very similar anticancer compounds (Imatinib and Masitinib) which only differ in one functional group: a Thiazole ring for Masitinib instead of a Piperazine ring for Imatinib. 
\begin{figure}[!htb]
\centering
\includegraphics[width=\linewidth]{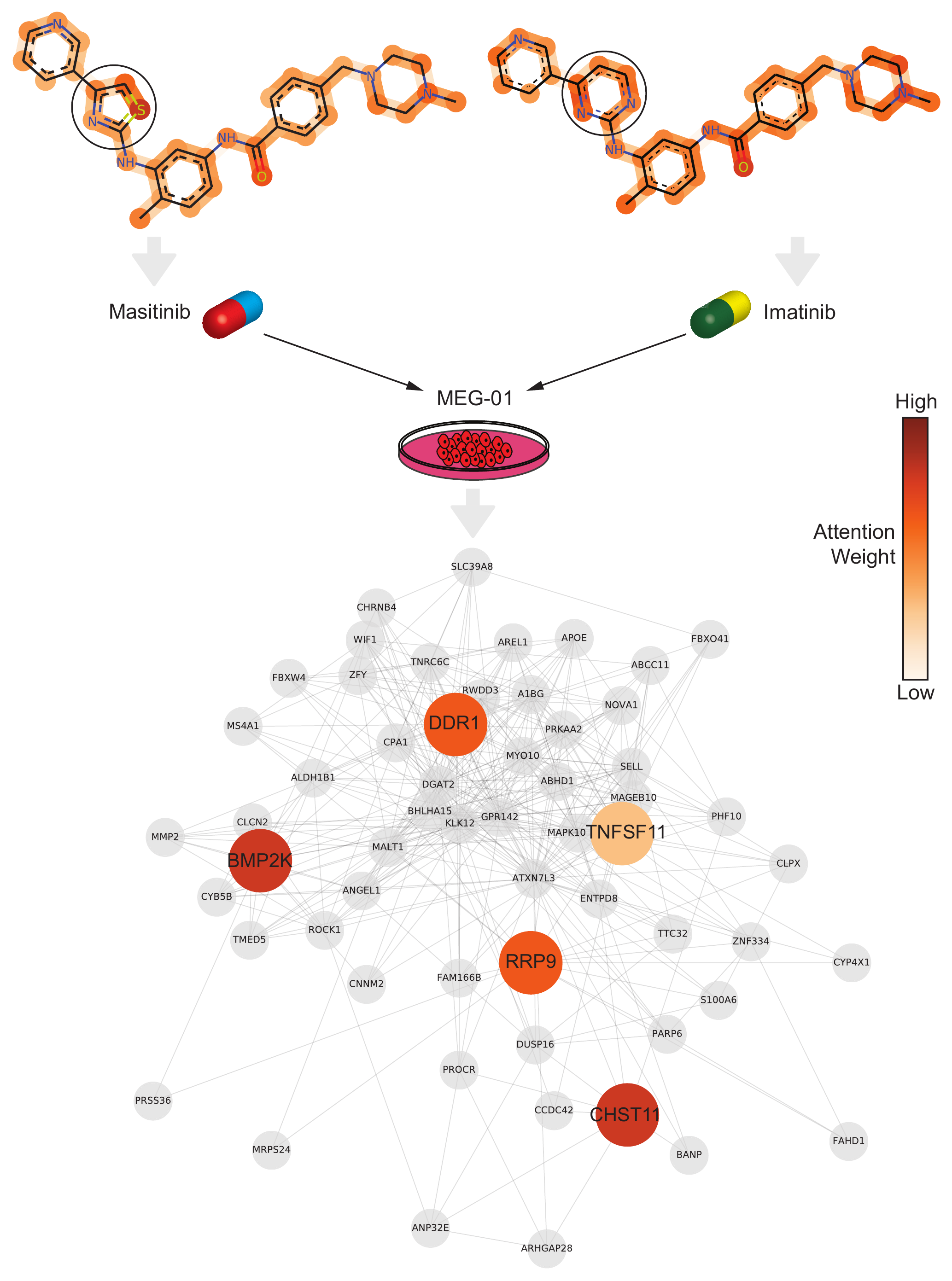}%
\caption{\textbf{Neural attention on molecules and genes.} The molecular attention maps on the top demonstrate how the model's attention is shifted when the Thiazole group is replaced by a Piperazine group. The change in attention across the two molecules is particularly concentrated around the affected rings, signifying that these functional groups play an important role in the mechanism of action for these tyrosine-kinase-inhibitors when they act on a chronic myelogenous leukemia (CML) cell line. The gene attention plot at the bottom depicts the most attended genes of the CML cell line, all of which can be linked to leukemia (details see text).}
\label{fig:attention_analysis}
\end{figure}
Both studied drugs are tyrosine-kinase-inhibitors that are predominantly applied in haematopoietic and lymphoid tissue. 
Generally, their IC50 values are highly correlated, particularly for their target cell lines ($\rho$ = 0.72).  
\autoref{fig:attention_analysis} depicts the attention over both molecules when paired with cell line MEG-01 (COSMIC ID 1295740, a type of chronic myelogenous leukemia). Leukemia is targeted quite successfully by both drugs, with Imatinib (IC50 = 81nM) being superior to Masitinib (223nM). 
%The model's predictions capture this trend solely from the different heterocyclic ring structures. 
Comparing the attention weights on both molecules depicted in \autoref{fig:attention_analysis} reveals that the attention weights on the affected functional groups (encircled) are drastically different in the two compounds whereas the remaining regions of the both molecules are primarily unaffected. The localized discrepancy in attention centered at the affected rings suggests that these substructures are of primary importance to the model in predicting the sensitivity of the MEG-01 cell line to Imatinib and Masitinib.

At the bottom of \autoref{fig:attention_analysis} the most attended genes of the studied leukemia cell line and their STRING protein neighborhoods are presented.
Interestingly, the DDR1 protein is a member of receptor tyrosine kinases (RTKs), the same group of cell membrane receptors that both Imatinib and Masitinib inhibit~\citep{kim2011ddr1}. DDR1 gene is highly expressed in various cancer types, such as in chronic lymphocytic leukaemia~\citep{barisione2017heterogeneous}. In addition, BMP2K gene has been recently shown to be implicated in chronic lymphocytic leukemia (CLL) ~\citep{pandzic2016transposon}, while CHST11 has long been known to be deregulated in CLL~\citep{schmidt2004deregulation}.
TNFSF11 encodes RANKL which is part of a prominent cancer signalling pathway~\citep{renema2016rank} and TNFSF11 has been reported to be the most overexpressed gene in a sample of $n$ = 129 acute lymphoblastic leukemia (ALL) patients ~\citep{heltemes2011ebf1}. RRP9 has been shown to be crucial in treating ALL~\citep{rainer2012research}). In conclusion, the prior knowledge from the cancer literature validate our findings and indicate that the genes that were given the highest attention weights by our model are indeed crucial players in the progression and treatment of leukemia. 

\section{Discussion}
\label{sec:discussion}

We presented an attention-based multimodal neural approach for explainable drug sensitivity prediction using a combination of 1) SMILES string encoding of drug compounds 2) transcriptomics of cancer cells and 3) intracellular interactions incorporated into a PPI network.
In an extensive comparative study of SMILES sequence encoders, We demonstrated that using the raw SMILES string of drug compounds, we were able to surpass the predictive performance reached by a baseline model utilizing Morgan fingerprints. In addition, we demonstrated that the attention-based SMILE encoder architectures, especially the newly proposed MCA, performed the best while producing results that were verifiably explainable. The validity of the drug attention has been corroborated by demonstrating its strong correlation with a well established structure similarity measure.
To further improve the explainabiltiy of our models, we devised a gene attention mechanism that acts on genetic profiles and focuses on genes that are most informative for IC50 prediction. We validated the correctness of the gene attention weights performing a pathway enrichment analysis over all the cell lines contained in GDSC and finding a significant enrichment of apoptotic processes.
In a case study on a leukemia cell line we have showcased how our model is able to focus on relevant compounds' structural elements and consider genes relevant for the disease of interest.

A key feature of our models was the strict training and evaluation strategy that set our work apart from previous art. In our strict model evaluation approach, cells and compounds were split in training, validation and test dataset before building the pairs, ensuring neither cells nor compounds in the validation or test datasets were ever seen by the trained model, thus depriving the model from a significant portion of available samples. Despite this unforgiving evaluation criterion, our best model (MCA) achieved an average standard deviation of 0.11 in predicting normalized IC50 values for unseen drug-cell pairs.
Furthermore, in a separate comparative study on the same dataset, this time with a lenient data split and model evaluation criterion, we demonstrated that our MCA model outperformed previously reported state-of-the-art results by achieving a RMSE of 0.89 and a R2 of 86\%.
We envision our attention-based approach to be of great utility in personalized medicine and de novo anticancer drug discovery where explainable prediction of drug sensitivity is paramount. Furthermore, having established a solid multimodal predictive model we have paved the way for future directions such as: 1) Drug repositioning applications as our model enables drug sensitivity prediction for any given drug-cell line pair. 2) Leveraging our model in combination with recent advances in small molecule generation using generative models~\citep{Kadurin2017, Blaschke2017} and reinforcement learning~\citep{popova2018deep} to design novel disease-specific, or even patient-specific compounds.
This opens up a scenario where personalized treatments and therapies can become a concrete option for patient care in cancer precision medicine.

\section{Availability of software and materials}
\label{ssec:availability}

The data in TFRecord format used in the benchmark studies conducted in this work can be downloaded at the following url: \url{https://ibm.biz/paccmann-data}.
Alternatively the reader can access the raw cell line data from \href{https://www.cancerrxgene.org/downloads}{GDSC}~\cite{iorio2016landscape} and the compound structural information from \texttt{\href{https://pubchem.ncbi.nlm.nih.gov/search/}{PubChem}}~\citep{kim2018pubchem} and the \texttt{\href{http://lincs.hms.harvard.edu/db/}{LINCS}} database.

The implementation of the models used in the benchmark is available in the form of a toolbox on Github a the following link: \url{https://github.com/drugilsberg/paccmann}.

Furthermore, the best MCA model has been deployed as a service on IBM Cloud.
Users can access the app and provide a compound in SMILES format to obtain a prediction of its efficacy in terms of IC50 on 970 cell lines from GDSC.
The results on drug sensitivity together with the top-attended genes can be examined in a tabular format and downloaded for further analysis.
The service is open access and users can register directly on the web application a the following url: \url{https://ibm.biz/paccmann-aas}.

% Leave this commented out - no acks within peer-review
\section*{Acknowledgments}
\label{ssec:acknowledgements}
The project leading to this publication has received funding from the European Union’s Horizon 2020 research and innovation programme under grant agreement No. 826121

\bibliographystyle{icml2019}
\begin{footnotesize}
\bibliography{main}
\end{footnotesize}

\end{document}